%% file: main.tex
\documentclass[10pt,twocolumn,letterpaper]{article}

\usepackage[pagenumbers]{iccv} 


\definecolor{iccvblue}{rgb}{0.21,0.49,0.74}
\usepackage[pagebackref,breaklinks,colorlinks,allcolors=iccvblue]{hyperref}
\usepackage{amssymb}
\usepackage{multirow} 

\def\paperID{5652} 
\def\confName{ICCV}
\def\confYear{2025}

\title{NukesFormers: Unpaired Hyperspectral Image Generation with Non-Uniform Domain Alignment}

\author{Jiaojiao~Li\\
{\tt\small jjli@xidian.edu.cn}
\and
Shiyao~Duan\\
{\tt\small syduan\_1@stu.xidian.edu.cn}
\and
Haitao~XU\\
{\tt\small xuhaitao@nssc.ac.cn}
\and
Rui~Song\\
{\tt\small rsong@xidian.edu.cn}
}

\begin{document}
\maketitle
\input{sec/0_abstract}    
\input{sec/1_intro}
\input{sec/2_formatting}
\input{sec/3_finalcopy}
{
    \small
    \bibliographystyle{ieeenat_fullname}
    \bibliography{main}
}


\end{document}

%% file: sec/0_abstract.tex
\begin{abstract}
The inherent difficulty in acquiring accurately co-registered RGB-hyperspectral image (HSI) pairs has significantly impeded the practical deployment of current data-driven Hyperspectral Image Generation (HIG) networks in engineering applications. Gleichzeitig, the ill-posed nature of the aligning constraints, compounded with the complexities of mining cross-domain features, also hinders the advancement of unpaired HIG (UnHIG) tasks.
In this paper, we conquer these challenges by modeling the UnHIG to range space interaction and compensations of null space through Range-Null Space Decomposition (RND) methodology. Specifically, the introduced contrastive learning effectively aligns the geometric and spectral distributions of unpaired data by building the interaction of range space, considering the consistent feature in degradation process. Following this, we map the frequency representations of dual-domain input and thoroughly mining the null space, like degraded and high-frequency components, through the proposed Non-uniform Kolmogorov-Arnold Networks. Extensive comparative experiments demonstrate that it establishes a new benchmark in UnHIG.
\end{abstract}

%% file: sec/1_intro.tex
\section{Introduction}
\label{sec:intro}

Hyperspectral imaging finds extensive applications in numerous fields, including drug analysis~\cite{mahdieh2022hyperspectral}, remote sensing detection \cite{scafutto2017hyperspectral} and image fusion \cite{dengbidirectional}. Nevertheless, due to the inherent limitations of imaging equipment, acquiring an adequate number of hyperspectral images (HSIs) with high quality, compared with RGBs or multispectral images (MSIs), is excessively time-consuming and prohibitively expensive. Furthermore, hardware enhancements tend to occur at a sluggish pace, thereby highlighting the pressing need to rely on software or algorithms to obtain a substantial number of HSIs. Currently, as one of the most efficient technologies for acquiring HSIs, Hyperspectral Image Generation (HIG) enables the generation with both fine spatial and spectral resolution from a single RGB or MSI\cite{zhao2020hierarchical,hu2022hdnet,cai2022mst++,yao2024specat}, which is modeled to learn inverse mapping from HSI.


\begin{figure}
\includegraphics[width=3.5in]{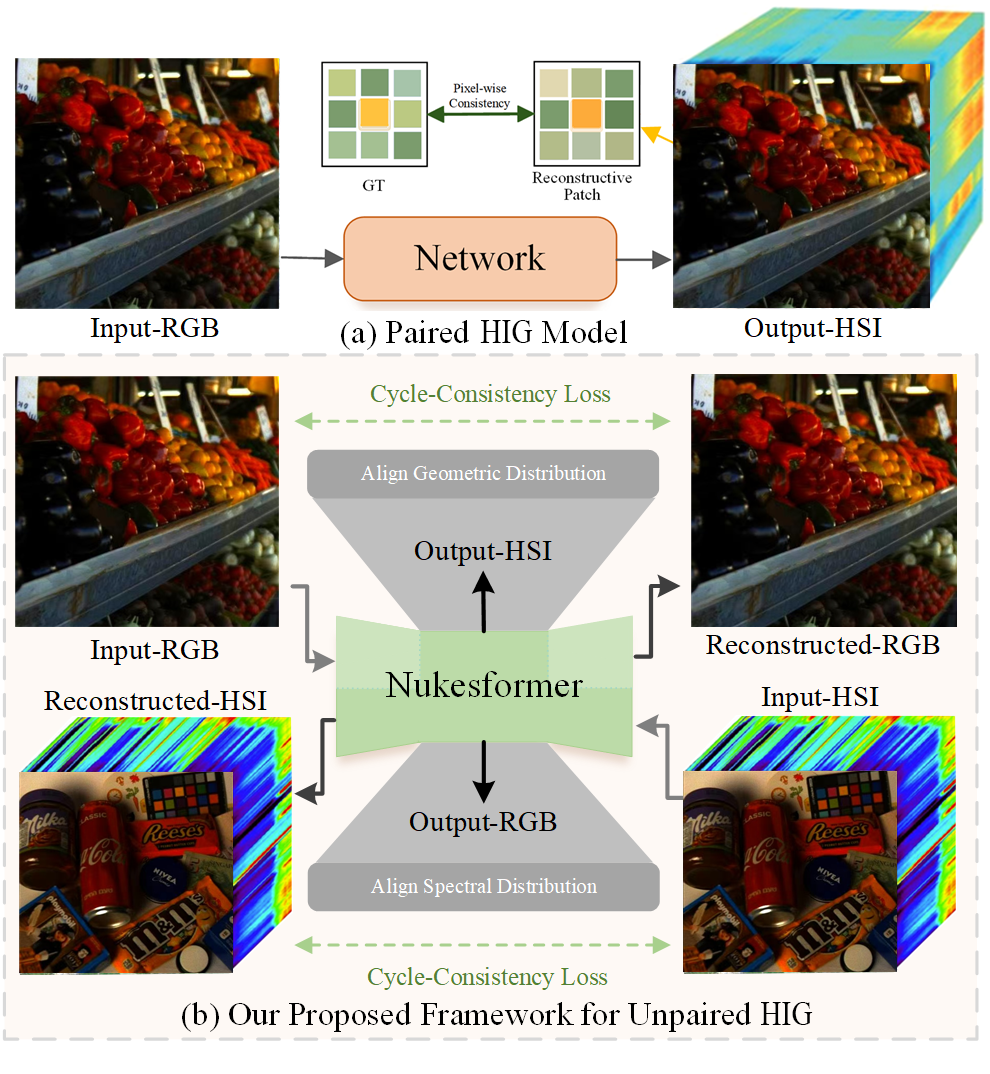}
\caption{(a) The conventional framework of paired HIG, which utilizes ground truth (GT) to build direct pixel-wise consistency. (b) Our proposed NukesFormer for unpaired HIG without GT, which establishes indirect constraint with cycle-consistency and dual-dimensional contrastive prior module (DCPM).}
\label{image1}
\end{figure}

Similar to other tasks, the data-driven learning strategy in HIG methods has recently witnessed a significant improvement in reconstruction performance \cite{cai2022coarse,dian2023spectral}, as depicted in Fig.\ref{image1}(a). Nonetheless, the reliability of fully-supervised HIG is often reliant on the amount and diversity of paired and registered RGB-HSI or MSI-HSI training sets, which are challenging to obtain in practical work yet.


To tackle these challenges, numerous studies have proposed training HIG models exploiting unsupervised \cite{zhu2021semantic,fubara2020rgb}, semi-supervised  \cite{hong2023decoupled}, and unpaired methods \cite{10198468,xie2024blind}, thereby mitigating the dependency on high-quality paired data with well co-registration. Among these approaches, unpaired HIG (UnHIG) has garnered increasing attention due to the prevalence of non-registered HSIs in practical applications \cite{qu2023unmixing,xie2024blind}. 
However, UnHIG is constrained by the absence of natural cross-domain interactions, leading to a serious non-uniform status between the corresponding information in dual domains. Additionally, the lack of human-annotated hints renders the valuable hyperspectral priors largely inaccessible during the training phase. As a result, critical degraded features, predominantly comprising high-frequency components, are either discarded or overwhelmed in existing approaches. Unfortunately, the challenges still seriously block the performance of UnHIG at present.

Drawing inspiration from the advances of Range-Null Space Decomposition (RND) methodologies in various tasks \cite{cheng2023null,yu2023range}, we explore the feasibility of leveraging RND to address the challenges of UnHIG. Specifically, we aim to decompose the inverse high-dimensional mapping into range-space interactions encompassing consistent features, and null-space compensations targeting degraded and high-frequency components. Within this framework, range-space interactions function as an indirect spectral and geometric constraint by encapsulating consistent or inconsistent information inherent to intra-domain and intra-scene relationships. On the other hand, the null-space compensation focuses on mining weak and attenuated high-frequency components of null space, thereby enhancing the overall reconstruction process.

In this paper, we propose a novel and lightweight framework for UnHIG by integrating cross-domain interactions and leveraging spectral characteristic information. Concretely, the unpaired data is systematically decomposed into continuous components termed the range space, and attenuated components referred to as the null space, via RND methodology. Subsequently, the framework employs a cycle-based architecture to capture intra-domain attributes, while utilizing dual-dimensional contrastive learning to aggregate continuous elements within the range space effectively. Moreover, building on Kolmogorov-Arnold Networks (KANs) \cite{liu2024kan,bresson2024kagnns}, a dedicated Non-Uniform Matrix Object-Aware Mechanism dynamically shifts the B-Spline function to enhance null-space compensation, facilitating high-dimensional fitting with robust guidance. Finally, extensive experimental evaluations demonstrate that the proposed method achieves state-of-the-art performance across various benchmarks.

The key contributions of this work are summarized as follows:

\begin{itemize}
\item We introduce a novel contrastive priori framework for UnHIG, which decomposes the generation process into null-space mining and range-space alignment, enabling superior high-frequency compensation and effective cross-domain interaction.

\item To adaptively extract high-frequency components obscured within the null space, we propose a Non-Uniform KANs (NukesFormers), which generates flexible and adaptive rational bases that precisely capture vulnerable yet crucial characteristic bands.

\item A dual-branch contrastive structure is further proposed that independently calibrates both geometric and spectral distributions to facilitate effective cross-domain interaction in range space, leveraging HSI and RGB or MSI from different scenes to build reliable constraint.
\end{itemize}

%% file: sec/2_formatting.tex
\section{Related Work}
\label{sec:related work}

\subsection{Hyperspectral Image Generation}
From the data-orientated perspective, HIG is categorized into three distinct categories: supervised \cite{cai2022mst++,li2020adaptive,hu2022hdnet}, prior-based unsupervised \cite{zhu2021semantic,li2023mformer}, and unpaired methods.

\renewcommand{\dblfloatpagefraction}{.9}
\begin{figure*}
\centering
\includegraphics[width=7.0in]{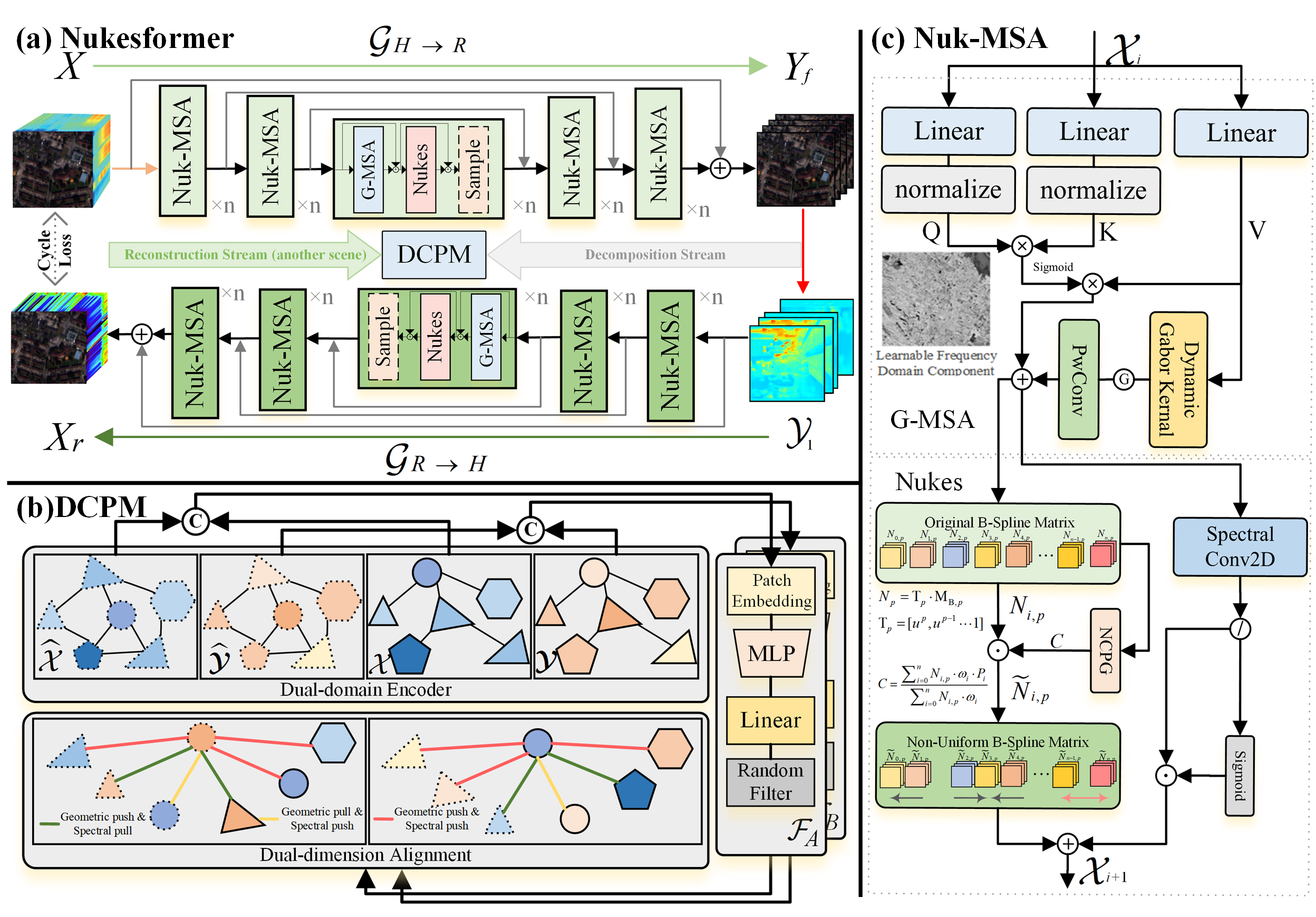}
\caption{The overall framework of proposed NukesFormer of decomposition stream, where the reconstruction stream in (a) derives from another NukesFormer, leveraging shared parameters. (b) DCPM utilizes the dual-dimensional contrastive prior to pull ($\xrightarrow{}\xleftarrow{}$) similar geometric and spectral distributions and push ($\xleftarrow{}\xrightarrow{}$) different components. (c) The non-uniform B-Spline matrix and G-MSA are dedicated to capturing multi-frequency information from spectral dimension.}
\label{image2}
\end{figure*}

Representing the most stringent data requirements and the most direct method of constraint construction, the supervised HIG utilizes many paired RGB-HSIs to construct the point-to-point constraint. Zhao et al. proposed that HRNet employs the residual dense block to obtain a larger receptive field and deploys the PixelShuffle to achieve feature fusion between layers \cite{zhao2020hierarchical}. Consequently, a self-supervised HIG framework was proposed \cite{zhu2021semantic}, which uses SRF and pre-trained semantic segmentation networks to construct point-level constraints, resulting in impressive results.

The aforementioned approaches neglect the vast collection of existing HSIs that have been captured and stored, due to the absence of essential prior, which implies that unpaired methods potentially reactivate the overlooked data to satisfy the requirements of industrial deployment.
\subsection{Unpaired Methods in Computer Vision}
The paradox between a vast amount of redundant unpaired data and the challenge of constructing appropriate constraints has consistently impeded the progress of computer vision, leading to the wastage of collected data \cite{liu2021unpaired,chen2022unpaired}. Initially, a Bayesian framework was constructed to establish the connection of two domains \cite{resales2003unsupervised}, utilizing Markov random fields in image-to-image translation. 

Additionally, the cycle-based \cite{zhu2017unpaired} approaches employ paired generators and discriminators to simulate domain information extraction, and introduce cycle consistency to achieve intuitive constraints \cite{engin2018cycle}. In deraining, contrastive prior is utilized to construct positive and negative sample pairs to extract and remove rain signals \cite{chen2022unpaired}. By utilizing contrastive learning methods to address patch information at the same position in images and retrieve information from distant parts, effective constraints are indirectly achieved \cite{park2020contrastive}.

In a similar vein, HIG tasks also strive to leverage ample unmatched HSI and RGB data \cite{zhu2021semantic}. This underscores the critical need to develop a model capable of achieving stable training and robust cross-scene interaction while relying solely on unpaired data.
\section{Method}
\label{sec:method}
In this section, we first present the MetaFormer \cite{yu2022metaformer} architecture equipped with Non-uniform KANs (NukesFormers) tailored for the UnHIG. In NukesFormers, the "Degradation" phase (Fig.\ref{image2}a) deploys the dynamic Gabor transformer to mine high-frequency components and the cascading Non-uniform KANs embedded Multi-head Self-attention (Nuk-MSA) to compensate for the attenuated null space in the RGB domain. Then, we introduce the dual-dimensional contrastive prior module (DCPM), designed to build the reliable constraint with range space.
\subsection{NukesFormers for UnHIG}

Unlike paired HIG, we employ an acquired HSI denoted $\textbf{X} \in \mathbb{R}^{C\times H \times W}$ along with an RGB of a different scene, represented as $\hat{\textbf{Y}} \in \mathbb{R}^{3\times H \times W}$, in UnHIG. Obviously, UnHIG cannot directly leverage GT to extract latent feature representations of HSI domain, but a large amount of unpaired RGB-HSIs still bury a unified degradation-reconstruction process as below:
\begin{equation}
\begin{aligned}
\label{deqn_ex1}
 \textbf{Y} = \textbf{D}\textbf{X}+\textbf{N}_{c}\quad \hat{\textbf{Y}} = \textbf{D}\hat{\textbf{X}}+\textbf{N}_{c}\\
\end{aligned}
\end{equation}
where $\textbf{D} \in \mathbb{R}^{3\times C}$ and $\textbf{N}_{c} \in \mathbb{R}^{3\times H \times W}$ are depicted as the SRF and camera noise. Furthermore, $\textbf{Y}$ means the RGB in the same scene as $\textbf{X}$.

Omitting $\textbf{N}_{c}$ considerations, the dual direction mapping process can be portrayed using the inverse operation (pseudo-inverse) of $\textbf{D}$ as $\textbf{D}^\dagger \in \mathbb{R}^{C\times 3}$, as illustrated in the RND methodology \cite{wang2023range}.
\begin{equation}
\begin{aligned}
\label{deqn_ex2}
 \textbf{X} \equiv \textbf{D}^\dagger\textbf{D}\textbf{X}+ (\textbf{I}-\textbf{D}^\dagger\textbf{D})\textbf{X}\\
  \hat{\textbf{Y}} \equiv \textbf{D}\textbf{D}^\dagger\hat{\textbf{Y}}+ (\textbf{I}-\textbf{D}\textbf{D}^\dagger)\hat{\textbf{Y}}\\
\end{aligned}
\end{equation}
which decomposes the HSIs into range space $\textbf{D}^\dagger\textbf{D}\textbf{X}$ and null space $(\textbf{I}-\textbf{D}^\dagger\textbf{D})\textbf{X}$.
Evidently, the cycle-based framework can be built by degradation-reconstruction operators $\textbf{D}^\dagger\textbf{D}(\cdot)$ as long as $\textbf{I}-\textbf{D}^\dagger\textbf{D} = \textbf{0}$, which implies that $\mathbf{D}$ must be orthogonal (this, however, is infeasible since $C > 3$). As a consequence, we redefine the equation as $ \textbf{min}||\textbf{X} - \textbf{D}^\dagger \textbf{D} \textbf{X}||_{2}^{2} , \textbf{s.t.} , ||(\textbf{I} - \textbf{D}^\dagger \textbf{D}) \textbf{X}||_{2}^{2} =\epsilon_{0}$, $\epsilon_{0}$ represents a value infinitesimally close to zero. Obviously, the above constraints can be simplified to $\textbf{min}||\textbf{D}^\dagger \textbf{D} \textbf{X}-\textbf{X}||_{2}^{2}$. Then the "Reconstruction" phase possesses a similar consequent, which establishes the cycle-based constraint as follows.
\begin{equation}
\begin{aligned}
\label{deqn_ex3}
  \mathcal{L}_{cyc}(\textbf{X}, \hat{\textbf{Y}})=&\mathbb{E}_{r\sim R}[||\mathcal{G}_{R\rightarrow H\rightarrow R}(\textbf{X})-\textbf{X}||_{2}^{2}] \\
 +&\mathbb{E}_{h\sim H}[||\mathcal{G}_{H\rightarrow R\rightarrow H}(\hat{\textbf{Y}})-\hat{\textbf{Y}}||_{2}^{2}]
\end{aligned}
\end{equation}
in which $\mathcal{G}_{R\rightarrow H\rightarrow R}$ and $\mathcal{G}_{H\rightarrow R\rightarrow H}$ represent the joint models of reconstruction-degradation ($\mathcal{G}_{R\rightarrow H}$-$\mathcal{G}_{H\rightarrow R}$) and degradation-reconstruction operations with NukesFormers, as depicted in Fig.\ref{image2}a.

\subsection{Non-uniform KANs Embedded Multi-head Self-attention}
To obtain the consistent feature of the RGB-HSI domain buried in unpaired data, we introduce the NukesFormer, accurately capturing geometric distributions and essential high-frequency information in null space. As depicted in Fig.\ref{image2}a, spectral degradation $\mathcal{G}_{H\rightarrow R}$ and reconstruction $\mathcal{G}_{R\rightarrow H}$ are individually modelled using paired NukesFormers, which are then integrated into dual domains through a shared parameter mechanism.
\begin{equation}
\begin{aligned}
\label{deqn_ex4}
  {\textbf{Y}}_{f} = \mathcal{G}_{H\rightarrow R}({\textbf{X}}|\theta_{H\rightarrow R}) \quad & {\textbf{X}}_{r} = \mathcal{G}_{R\rightarrow H}({\textbf{Y}}_{f}|\theta_{R\rightarrow H}) \\
  \hat{\textbf{X}}_{f} = \mathcal{G}_{R\rightarrow H}(\hat{\textbf{Y}}|\theta_{R\rightarrow H}) \quad & \hat{\textbf{Y}}_{r} = \mathcal{G}_{H\rightarrow R}(\hat{\textbf{X}}_{f}|\theta_{H\rightarrow R})\\
\end{aligned}
\end{equation}
in which $\theta_{R\rightarrow H}$ and $\theta_{H\rightarrow R}$ represent the learnable parameter. In the "Degradation" phase, acquired HSIs $\textbf{X}$ undergo a modelled degradation process to produce the corresponding fake-RGB ${\textbf{Y}}_{f}$. Subsequently, a NukesFormer simulates the reconstruction process, converting the fake RGB back to a recovery-HSI ${\textbf{X}}_{r}$. In the "Reconstruction" phase, the RGB of another scene experiences a similar process.

\noindent \textbf{Nuk-MSA.}
However, one can anticipate that $\mathcal{G}_{H\rightarrow R}$ might yield high-frequency components of HSI domain that are overly degraded, due to the abstraction of intermediate feature as the network deepens. Initially, we sample the RGB or HSI to obtain the original feature. 
\begin{equation}
\begin{aligned}
\label{deqn_ex4-1}
  \mathcal{X}_{1} = Conv_{1\times1}(\textbf{X}) \quad\mathcal{Y}_{1} = Conv_{1\times1}(\textbf{Y}_{f})
\end{aligned}
\end{equation}
Subsequently, these features undergo a progressive refinement process via a series of cascaded Nuk-MSA.
\begin{equation}
\begin{aligned}
\label{deqn_ex4-2}
  \mathcal{X}_{i+1} = \textbf{M}_{DGK} (\mathcal{X}_{j}|\theta_{en,i+1}^{H\rightarrow R})&\quad\mathcal{Y}_{i+1} = \textbf{M}_{DGK} (\mathcal{Y}_{i}|\theta_{en,i+1}^{R\rightarrow H})\\
  \mathcal{X}_{j+1} = \textbf{M}_{DGK}(Map(C&oncat(\mathcal{X}_{j}, \mathcal{X}_{i}))|\theta_{de,j+1}^{H\rightarrow R})\\
  \mathcal{Y}_{j+1} = \textbf{M}_{DGK}(Map(C&oncat(\mathcal{Y}_{j}, \mathcal{Y}_{i}))|\theta_{de,j+1}^{R\rightarrow H})\\
\end{aligned}
\end{equation}
$\mathcal{X}_{i}, \mathcal{Y}_{i}$ are the intermediate feature of every encoder in NukesFormer, and $\mathcal{X}_{j}, \mathcal{Y}_{j}$ represent the feature of the corresponding level in decoder.
\begin{equation}
\begin{aligned}
\label{deqn_ex4-3}
  \textbf{Y}_{f} = \mathcal{X}_{F} + \mathcal{X}_{1}\quad \textbf{X}_{r} = \mathcal{Y}_{F} + \mathcal{Y}_{1}
\end{aligned}
\end{equation}
Finally, a skip connect is utilized to gain the $\textbf{Y}_{f}$ and $\textbf{X}_{r}$.

\noindent \textbf{Gabor Kernel based Multi-head Self-attention (G-MSA).}
In contrast to frequency domain transformations that rely on FFT or similar techniques \cite{huang2023adaptive,li2023pixel,chi2020fast,li2020fourier}, the utilization of Gabor kernels enables the model to focus more on local and needed frequency intervals, which effectively mitigates the issue of high-frequency information being overshadowed.

Specifically, we employ the spectral multi-head self-attention (MSA) to generate the requisite query $\textbf{Q}\in\mathbb{R}^{HW\times C}$, key $\textbf{K}^{T}\in\mathbb{R}^{C\times HW}$, and value $\textbf{V}$ through linear operations, thus obtaining global spectral features.
\begin{equation}
\begin{aligned}
\label{deqn_ex5}
  \textbf{Q}, \textbf{K}, \textbf{V} = Linear(\mathcal{X}_{i}|\theta_{Q}, \theta_{K}, \theta_{V}) \\
  \textbf{A} = Softmax(\textbf{K}^{T}\textbf{Q})\quad \mathcal{X}_{i, msa} = \textbf{V}\textbf{A}\\
\end{aligned}
\end{equation}
in which $\theta_{Q}, \theta_{K}$ and $\theta_{V}$ represent the learnable parameters.

To mine deeper into high-frequency feature in varying scenes, a scene-adaptive Gabor kernel is embedded by dynamic convolution layer \cite{nam2022frequency,chen2020dynamic,jia2016dynamic} to facilitate feature generation from various directions and frequency domain intervals. Notably, the range of intervals is contingent upon the specific objects under consideration.
\begin{equation}
\begin{aligned}
\label{deqn_ex6}
  g_{m}(x, y|f_{m}) = exp(-\frac{x_{\theta}^{2}+(y_{\theta}^{2}/f_{m})^{2}}{2\sigma^{2}})
\end{aligned}
\end{equation}
where $f_{m}$ is the $\textit{m-th}$ dynamic frequency parameter to control the interval of the Gabor kernel, $x_{\theta}=xcos(\theta)+ysin(\theta)$ and $y_{\theta}=-xsin(\theta)+ycos(\theta)$. And $x$ and $y$ mean the horizontal and vertical position coordinates in spatial dimension.

As depicted in Fig.\ref{image2}c, the Gabor kernels are embedded into a dynamic depthwise convolution to lightweigh. Subsequently, the multiple windows captured are mapping to one learnable frequency domain component for every channel.
\begin{equation}
\begin{aligned}
\label{deqn_ex7}
\textbf{V}_{f,m} = \textbf{V} * g_{m}&(x, y) \quad  \textbf{V}_{f}=\{\textbf{V}_{f,1}, \textbf{V}_{f,2}, ...\}\\
\mathcal{X}_{i, fry} &= Conv(GELU(\textbf{V}_{f}))\\
\end{aligned}
\end{equation}
by which the frequency value $\textbf{V}_{f}\in \mathbb{R}^{H\times W\times CM}$ are mapping to $\mathcal{X}_{i, fry} \in \mathbb{R}^{H \times W\times C}$.
After that, the final feature in G-MSA is obtained with the element-wise addition operation.
\begin{equation}
\begin{aligned}
\label{deqn_ex8}
\mathcal{X}_{i, f} = \mathcal{X}_{i, fry} + \mathcal{X}_{i, msa}
\end{aligned}
\end{equation}

\noindent \textbf{Non-uniform KANs Networks (Nukes)}
In the conventional MetaFormer architecture, multi-head self-attention (MSA) generates multiple attention maps that are typically integrated via concatenated MLP or CNN-based feed-forward networks (FFN). However, they often neglect the intricate nature of spectral features and the capacity to capture high-dimensional characteristics critical for hyperspectral tasks. Therefore, UnHIG imposes heightened requirements on capacity of building high-dimensional mapping in FFN.

In response to the challenges of high-dimensional spectral fitting in UnHIG, we introduce a lightweight and interpretable framework, Nukes. Leveraging advanced KANs, that streamline the traditionally redundant linear weights and nonlinear activation functions into a compact set of initial learnable parameters $\omega$. These parameters are used within a recursive approach to generate multi-level B-spline basis functions $\textbf{N}_{i,p}$.

\begin{equation}
\label{deqn_Bspline}
\textbf{N}_{i,0}(x)= \begin{cases}
1,\quad & x_{i}\leq x \textless x_{i+1}\\
0,\quad & otherwise
\end{cases} 
\end{equation}

\begin{equation}
\begin{aligned}
\label{deqn_Bspline2}
\textbf{N}_{i,p}(x) &= \frac{x-x_{i}}{x_{i+p}-x_{i}}\textbf{N}_{i,p-1}(x) \\
&+\frac{x_{i+p+1}-x}{x_{i+p+1}-x_{i+1}}\textbf{N}_{i+1,p-1}(x)
\end{aligned}
\end{equation}

While KANs offer a powerful recursive framework, they are inherently GPU-unfriendly due to recursive dependencies. To enhance parallel computation and efficiency, we reformulate the recursive process into matrix form as Eq. \ref{deqn_Bspline3}, achieving significantly improved computational performance on GPU architectures.

\begin{equation}
\begin{aligned}
\label{deqn_Bspline3}
\textbf{N}_{p} &= \textbf{T}_{p}\textbf{M}_{B, p} \\
\textbf{T}_{p} &= [\textbf{u}^{p}, \textbf{u}^{p-1}, \cdots, 1]
\end{aligned}
\end{equation}
where $\textbf{N}_{p}$, $\textbf{T}_{p}$ and $\textbf{M}_{B, p}$ represent the weight matrix of control points, multi-stage index matrix and B-spline basis matrix constructed by learnable parameters.

However, capturing essential high-dimensional spectral features in UnHIG remains challenging due to limited nonlinear representation and uniform reference point placement along the coordinate axis. To address this, we propose a Non-Uniform Control Point Generator (NCPG), which dynamically adjusts control points’ relative attraction and exclusion forces within the Nukes curve. During training, control points $\textbf{P}_{i}$ are adaptively concentrated on the most salient regions of the high-dimensional spectral features.
\begin{equation}
\begin{aligned}
\label{deqn_NURBs}
 \textbf{Nuk}(x)=\frac{\sum_{i=0}^{n}\textbf{N}_{i,p}(x)\cdot \omega_{i}\cdot \textbf{P}_{i}}{\sum_{i=0}^{n}\textbf{N}_{i,p}(x)\cdot \omega_{i}}
\end{aligned}
\end{equation}
in which $\textbf{Nuk}(x)$ is the non-uniform B-spline curve.

Finally, the another branch with spectral CNN extract the low-frequency information to complement the geometric features as Eq. \ref{deqn_ex9}.
\begin{equation}
\begin{aligned}
\label{deqn_ex9}
&[\mathcal{X}_{i, up}, \mathcal{X}_{i, dw}] = Chunk(Spec\text{-}Conv(\mathcal{X}_{i, f})) \\
&\mathcal{X}_{i+1} = \alpha\textbf{Nuk}(\mathcal{X}_{i, f})+\mathcal{X}_{i, up}\cdot sigmoid(\mathcal{X}_{i, dw})\\
\end{aligned}
\end{equation}
where $\alpha$ is the learnable ratio.

In summary, Nukes aims to minimize parameter usage while maintaining interpretability, allowing it to adaptively fit and learn critical high-dimensional features and mapping relationships in hyperspectral tasks.

\renewcommand{\dblfloatpagefraction}{.99}
\begin{figure*}[!t]
\centering
\includegraphics[width=1.00\linewidth]{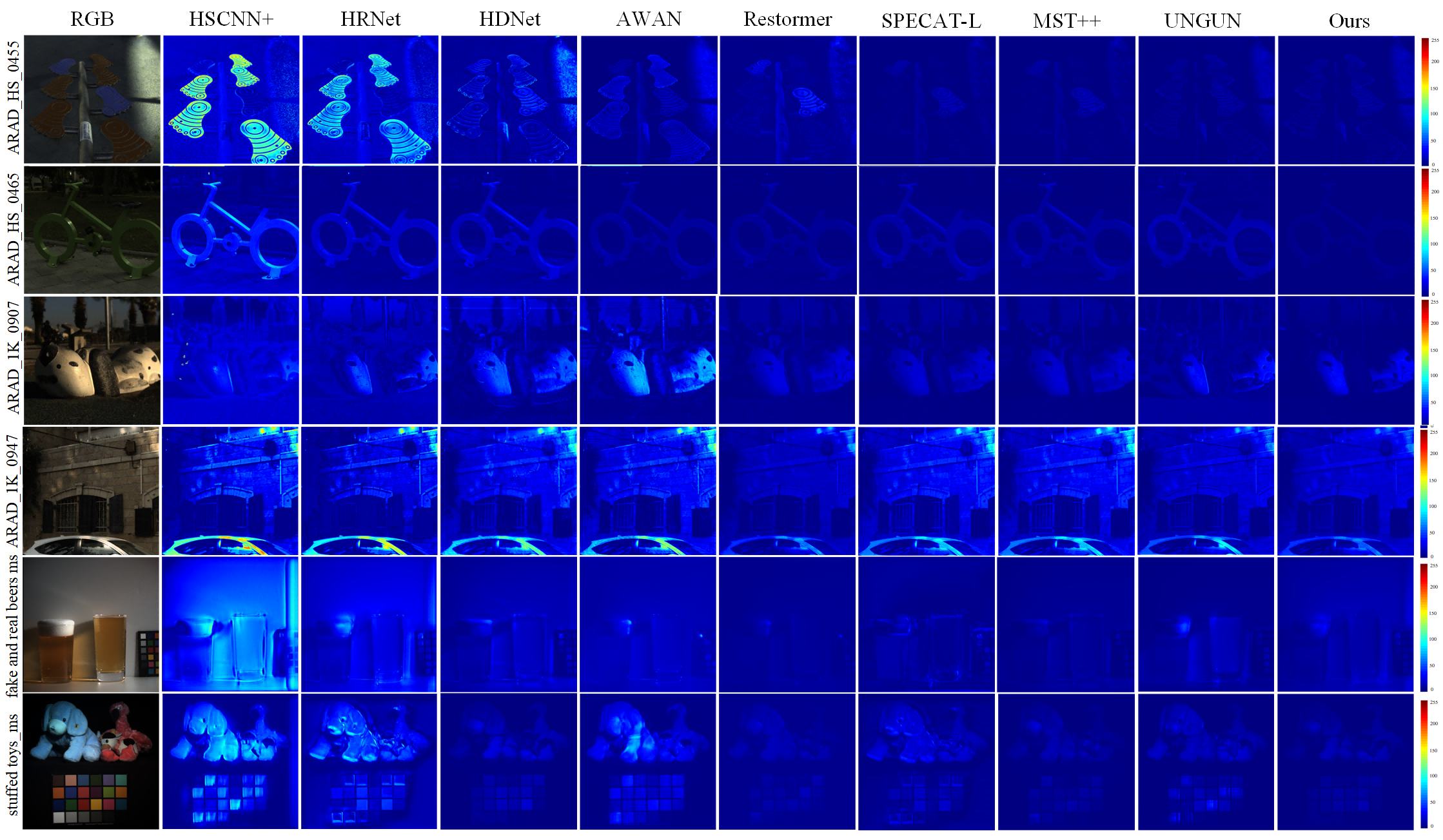}
\caption{\textbf{Visual Comparison Results}. Line 1-2: the RMSE error map of 20-th band on three distinct validation images in NTIRE 2020 Clean. Line 3-4: the RMSE error map of 21-st band on three distinct validation images in NTIRE 2022. Line 5-6: the RMSE error map of 5-th band on two distinct validation images in CAVE.}
\label{image3}
\end{figure*}
\subsection{Dual-Dimensional Contrastive Prior Module}
In pursuit of stable constraint, the DCPM is employed to facilitate interaction between the two domains, by building the relationship between RGB and HSI of distinct scenes with range space. 
However, distinguishing from existing methods \cite{chen2022unpaired,chen2020simple}, DCPM introduces dual-dimensional contrastive priors, which are dedicated to resolving with concurrently handling mutual information from both domains, which are embedded with significantly different feature distribution.

Consequently, DCPM incorporates two auxiliary encoders ($\mathcal{F}_{A}(\cdot)$ and $\mathcal{F}_{B}(\cdot)$) that map the output of final stage's Nuk-MSA to the spectral and geometric features of latent spaces, in which the auxiliary encoders are constituted by multilayer perceptron (MLP) with the activation function. Illustrating with the "Degradation" phase, the features extracted from both domains are divided into patches. Then, a single patch from one domain is designated as the query code $\textbf{f}$. Subsequent patches from both domains are then partitioned into positive $\textbf{f}^{+}$ and negative codes $\textbf{f}^{-}$. It's noteworthy that patches from both domains are utilized in distinguishing positive from negative samples, marking a deviation from previous methods \cite{park2020contrastive}. After filters are applied to randomly extract relevant patches, the resultant samples undergo spectral contrastive learning.

\begin{equation}
\begin{aligned}
\label{deqn_ex10}
 &\mathcal{L}_{spec}(\mathcal{G}_{R\rightarrow H}, \mathcal{G}_{H\rightarrow R})= \\
 &\mathbb{E}_{h\sim H, r\sim R}\left[-\mathrm{log}\frac{\mathrm{sam}(\mathbf{f}, \mathbf{f}^{+})}{\mathrm{sam}(\mathbf{f}, \mathbf{f}^{+}) +\sum^{N}_{i=1}\mathrm{sam}(\mathbf{f}, \mathbf{f}^{-}_{i})}\right] \\
\end{aligned}
\end{equation}
where the $\mathrm{sam}$ represents the spectral angle mapper (SAM) to quantify spectral similarity between quary code 
and $\mathbf{f}^{+}$ or $\mathbf{f}^{-}$. Additionally, we also align the geometric distribution in the "Reconstruction" phase with ralative geometric position. 
\begin{equation}
\begin{aligned}
\label{deqn_ex11}
 &\mathcal{L}_{geo}(\mathcal{G}_{R\rightarrow H}, \mathcal{G}_{H\rightarrow R})= \\
 &\mathbb{E}_{h\sim H, r\sim R}\left[-\mathrm{log}\frac{\mathrm{sim}(\mathbf{f}, \mathbf{f}^{+})}{\mathrm{sim}(\mathbf{f}, \mathbf{f}^{+}) +\sum^{N}_{i=1}\mathrm{sim}(\mathbf{f}, \mathbf{f}^{-}_{i})}\right] \\
\end{aligned}
\end{equation}
in which $\mathrm{sim}(u^{T}, v) = \mathrm{exp}(\frac{u^{T} v}{||u|| ||v|| \tau})$ represent the cosine similarity function and $\tau$ is the temperature parameter.

\renewcommand\arraystretch{1.5}
\begin{table*}[!t]
\centering
\caption{The experimental results of NTIRE 2020 Clean, NTIRE2022, and CAVE datasets in UnHIG. The best and second best results are $\mathbf{highlighted}$ and \underline{underlined} as below. The existing unpaired framework, UnGUN, is divided with the line. * represents the SRF is unavailable in this dataset.}
    \label{DATAresults}
\begin{tabular}{c|cccc|cccc|ccc}
\hline
\multirow{2}{*}{\textbf{Method}}  & \multicolumn{4}{c|}{\textbf{NTIRE 2020 Clean}}                                                                                          & \multicolumn{4}{c|}{\textbf{NTIRE 2022}}                                                                                                      & \multicolumn{3}{c}{\textbf{CAVE}}                                    \\ \cline{2-12} 
                        & \multicolumn{1}{c}{RMSE}             & \multicolumn{1}{c}{MRAE}            & \multicolumn{1}{c}{PSNR}           & SSIM      & \multicolumn{1}{c}{RMSE}             & \multicolumn{1}{c}{MRAE}            & \multicolumn{1}{c}{PSNR}           & SSIM            & \multicolumn{1}{c}{RMSE}       & \multicolumn{1}{c}{PSNR}       & SSIM \\ \hline
HSCNN+                  & \multicolumn{1}{c}{0.0976}                 & \multicolumn{1}{c}{0.3312}                & \multicolumn{1}{c}{19.53}               &   \multicolumn{1}{c|}{0.7002}  & \multicolumn{1}{c}{0.1124}                 & \multicolumn{1}{c}{1.2473}                & \multicolumn{1}{c}{20.68}               &     \multicolumn{1}{c|}{0.4107}  & \multicolumn{1}{c}{0.0739}          & \multicolumn{1}{c}{22.86}          &\multicolumn{1}{c}{0.7123} \\
HRNet                   & \multicolumn{1}{c}{0.0597}                 & \multicolumn{1}{c}{0.2346}                & \multicolumn{1}{c}{24.93}               &  \multicolumn{1}{c|}{0.8329}  & \multicolumn{1}{c}{0.0991}                 & \multicolumn{1}{c}{0.9019}                & \multicolumn{1}{c}{22.10}               &       \multicolumn{1}{c|}{0.5282}    & \multicolumn{1}{c}{0.0690}          & \multicolumn{1}{c}{24.09}           & \multicolumn{1}{c}{0.7392}\\
HDNet                   & \multicolumn{1}{c}{0.0467}                 & \multicolumn{1}{c}{0.1763}                & \multicolumn{1}{c}{27.57}               &  \multicolumn{1}{c|}{0.8862} & \multicolumn{1}{c}{0.0646}                 & \multicolumn{1}{c}{0.7537}                & \multicolumn{1}{c}{25.22}               &       \multicolumn{1}{c|}{0.8049}   & \multicolumn{1}{c}{0.0402}          & \multicolumn{1}{c}{28.14}           & \multicolumn{1}{c}{0.8869}  \\
AWAN                    & \multicolumn{1}{c}{0.0475}                 & \multicolumn{1}{c}{0.1897}                & \multicolumn{1}{c}{26.98}               &  \multicolumn{1}{c|}{0.8711}  & \multicolumn{1}{c}{0.0778}                 & \multicolumn{1}{c}{0.7916}                & \multicolumn{1}{c}{23.69}               &  \multicolumn{1}{c|}{0.7143}  & \multicolumn{1}{c}{0.0528}          & \multicolumn{1}{c}{26.02}           & \multicolumn{1}{c}{0.7951}  \\ 
Restormer               & \multicolumn{1}{c}{0.0389}                 & \multicolumn{1}{c}{\underline{0.1523}}                & \multicolumn{1}{c}{29.20}               &  \multicolumn{1}{c|}{0.9082} & \multicolumn{1}{c}{\underline{0.0573}}                 & \multicolumn{1}{c}{0.7285}                & \multicolumn{1}{c}{\underline{26.69}}               &    \multicolumn{1}{c|}{\underline{0.8350}}   & \multicolumn{1}{c}{\underline{0.0399}}          & \multicolumn{1}{c}{\underline{28.75}}           & \multicolumn{1}{c}{0.9098}      \\ 
SPECAT-L             & \multicolumn{1}{c}{0.0408}                 & \multicolumn{1}{c}{0.1617}                & \multicolumn{1}{c}{28.89}               &  \multicolumn{1}{c|}{0.9102} & \multicolumn{1}{c}{0.0637}                 & \multicolumn{1}{c}{0.7528}                & \multicolumn{1}{c}{26.22}               &    \multicolumn{1}{c|}{0.8288}   & \multicolumn{1}{c}{0.0412}          & \multicolumn{1}{c}{28.30}           & \multicolumn{1}{c}{\underline{0.9111}}      \\ 
MST++                   & \multicolumn{1}{c}{{0.0384}}    & \multicolumn{1}{c}{0.1542}          & \multicolumn{1}{c}{\underline{29.24}}               &   {0.9113}   & \multicolumn{1}{c}{0.0609}                 & \multicolumn{1}{c}{\underline{0.7002}}                & \multicolumn{1}{c}{26.46}               &  \multicolumn{1}{c|}{0.8264}   & \multicolumn{1}{c}{0.0405} &\multicolumn{1}{c}{28.28}                    &  \multicolumn{1}{c}{0.9107} \\ \hline
UnGUN             & \multicolumn{1}{c}{\underline{0.0362}}                 & \multicolumn{1}{c}{0.1675}                & \multicolumn{1}{c}{28.63}               &  \multicolumn{1}{c|}{\underline{0.9248}} & \multicolumn{1}{c}{0.0793*}                 & \multicolumn{1}{c}{0.7832*}                & \multicolumn{1}{c}{25.04*}               &    \multicolumn{1}{c|}{0.7942*}   & \multicolumn{1}{c}{0.0477}          & \multicolumn{1}{c}{27.93}           & \multicolumn{1}{c}{0.8832}      \\ \hline
Ours                    & \multicolumn{1}{c}{\textbf{0.0312}} & \multicolumn{1}{c}{\textbf{0.1476}} & \multicolumn{1}{c}{\textbf{30.55}} & \textbf{0.9420} & \multicolumn{1}{c}{\textbf{0.0528}} & \multicolumn{1}{c}{\textbf{0.4829}} & \multicolumn{1}{c}{\textbf{27.17}} & \textbf{0.8526} & \multicolumn{1}{c}{\textbf{0.0386}} & \multicolumn{1}{c}{\textbf{29.57}}  &  \multicolumn{1}{c}{\textbf{0.9275}}  \\ \hline
\end{tabular}
\label{tabel1}
\end{table*}

\subsection{More Details of Framework}
Given the absence of paired RGB-HSIs, constructing pixel-wise constraints directly becomes infeasible. As a result, NukesFormers are compelled to build more implicit or indirect constraints to optimize the $\mathcal{G}_{R\rightarrow H}$ and $\mathcal{G}_{H\rightarrow R}$.

\noindent \textbf{Adversarial Loss.}
Obviously, the absence of direct constraints on the intermediate outputs, $\hat{\textbf{X}}_{f}$ and $\textbf{Y}_{f}$, causes the reconstruction results of NukesFormers to resemble features of the latent space more than HSIs. Recognizing this, we introduce the adversarial loss. By feeding both the reconstructed and real dual-domain images from the dataset into  $\mathcal{L}_{adv}$, we aim to ensure that the final results are HSIs embodying the right spatial and spectral distribution, not merely feature maps.
\begin{equation}
\begin{aligned}
\label{deqn_ex12}
  \mathcal{L}_{adv}^{hsi}&=\mathbb{E}_{rh\sim H}logD_{H}(\textbf{X})\\
  &+\mathbb{E}_{fh\sim H}[log(1-D_{H}(\hat{\textbf{X}}_{f}))]\\
\end{aligned}
\end{equation}
Similar to other cycle-based methods \cite{chen2022unpaired,zhu2019singe}, $\mathcal{L}_{adv}^{hsi}$ makes the reconstructed-HSI similar to real distribution in HSI domain. And the dual-domain adversarial loss is calculated by $\mathcal{L}_{adv} = \mathcal{L}_{adv}^{hsi}+\mathcal{L}_{adv}^{rgb}$.

\noindent \textbf{Non-degraded Loss.}
Furthermore, inspired by the non-degradable properties of the spectrum, we introduced non-degraded loss to independently constrain $\mathcal{G}_{R\rightarrow H}$ and $\mathcal{G}_{H\rightarrow R}$. Specifically, when an RGB image undergoes the process represented by $\mathcal{G}_{H\rightarrow R}$, which denotes spectral degradation, its spectral features remain aligned, and the converse is also true. Consequently, $\mathcal{L}_{nde}$ is used to separately ensure their functional reliability.
\begin{equation}
\begin{aligned}
\label{deqn_ex13}
  \mathcal{L}_{nde}=&\mathbb{E}_{r\sim R}[||\hat{\mathcal{G}}_{R\rightarrow H}(\textbf{X})-\textbf{X}||_{2}^{2}] \\
 +&\mathbb{E}_{h\sim H}[||\hat{\mathcal{G}}_{H\rightarrow R}(\hat{\textbf{Y}})-\hat{\textbf{Y}}||_{2}^{2}]
\end{aligned}
\end{equation}
Notes that $\hat{\mathcal{G}}_{R\rightarrow H}$ and $\hat{\mathcal{G}}_{H\rightarrow R}$ represent the $\mathcal{G}_{R\rightarrow H}$ and $\mathcal{G}_{H\rightarrow R}$ removed the first up and downsampling operator.

To sum up, the $\mathcal{L}_{total}$ is defined as below:
\begin{equation}
\begin{aligned}
\label{deqn_ex14}
  \mathcal{L}_{total} =& \lambda_{1}*\mathcal{L}_{cyc} + \lambda_{2}*\mathcal{L}_{nde}\\ +& \lambda_{3}*\mathcal{L}_{adv} +\lambda_{4}*\mathcal{L}_{spec} + 
  \lambda_{5}*\mathcal{L}_{geo} \\
\end{aligned}
\end{equation}
in which the $\lambda_{i}$ represented as hyper-parameters. The specific setting in our experiment is provided in next session.

%% file: sec/3_finalcopy.tex
\section{Experiments}
\label{sec:experiments}

\subsection{Experimental Setting}

In comparison experiments, we select three benchmark dataset, consisting of NTIRE 2020 Clean Track \cite{arad2020ntire}, NTIRE 2022 \cite{arad2022ntire} and CAVE \cite{yasuma2010generalized}. 

To demonstrate the efficiency and adaptability of our proposed framework, we integrate HSCNN+ \cite{shi2018hscnn+}, HRNet \cite{zhao2020hierarchical}, HDNet \cite{hu2022hdnet}, AWAN \cite{li2020adaptive}, Restormer\cite{zamir2022restormer}, and MST++ \cite{cai2022mst++} as feature extraction units using existing state-of-the-art methods. In addition, another unpaired method UnGUN \cite{qu2023unmixing}.

During the experiment, the optimal learning rate was adopted for all the experiments. Based on this foundation, \textbf{Ours} adopts $n=\{1,2,4,2,1\}$. Furthermore, to balance the learning process between the generator and discriminator, the default loss weights were set as $\lambda_{1}=\lambda_{3}=1.0$, $\lambda_{2}=0.5$ and $\lambda_{4}=\lambda_{5}=0.25$.

Regarding the experimental environment, our work was conducted on the Pytorch platform utilizing the NVIDIA 4090 GPU. The Adam optimizer and cosine learning rate, consisting of $\beta_{1}$ = 0.9, $\beta_{2}$ = 0.999 and initial learning rate $lr_{init}$ = 0.0002, were employed as the primary optimization mechanisms during the experiment. Additionally, our methods were trained with $\mathrm{batch\_size=1}$ and $\mathrm{epoch=200}$.
\subsection{Performance Evaluation}
In this section, we segment the experimentation into two distinct parts: firstly, to validate the reliability and robustness of the NukesFormers against existing models, and secondly, to assess the UnHIG performance with ablation experiments. The results offer a holistic testament to the efficacy of the NukesFormers under the UnHIG paradigm, underscoring its superior performance in practical applications.
\subsubsection{Comparative Experiments}
Notably, CNN-based approaches often require a vast number of parameters, which implies they need meticulous hyper-parameter tuning when training in unpaired task. Comparatively, the Transformer-based method, as indicated in Fig.\ref{image2}, presents a darker overall error map, indicating its superior reconstruction effect.

\renewcommand\arraystretch{0.8}
\begin{table}[ht]
\caption{The comparison experimental results of Three datasets about four metrics and parameters. The best and second best results are \textbf{highlighted} and \underline{underline}.}
\resizebox{\linewidth}{!}{
\begin{tabular}{c|cc|c|c|c}
\hline
\multirow{2}{*}{*} & \multicolumn{2}{c|}{Params(M)}       & \begin{tabular}[c]{@{}c@{}}NTIRE\\ 2020\end{tabular} & \begin{tabular}[c]{@{}c@{}}NTIRE\\ 2022\end{tabular} & \multicolumn{1}{c}{\begin{tabular}[c]{@{}c@{}}CAVE\\ \end{tabular}}                             \\ \cline{2-6} 
                   & \multicolumn{1}{c|}{Train} & Infer & SAM                                                  & SAM                                                  & \multicolumn{1}{c}{SAM} \\ \hline
HSCNN+             & \multicolumn{1}{c|}{6.09}      &\multicolumn{1}{c|}{2.71}      &\multicolumn{1}{c|}{8.35}                                                      &\multicolumn{1}{c|}{11.33}                                                      & \multicolumn{1}{c}{19.35}      \\
HRNet              & \multicolumn{1}{c|}{44.41}      &  \multicolumn{1}{c|}{41.16}    &  \multicolumn{1}{c|}{6.01}      &   \multicolumn{1}{c|}{10.70}    & \multicolumn{1}{c}{16.58}      \\
HDNet              & \multicolumn{1}{c|}{6.15}      &  \multicolumn{1}{c|}{2.67}    &  \multicolumn{1}{c|}{5.89}  &  \multicolumn{1}{c|}{9.93}            & \multicolumn{1}{c}{13.55}             \\
AWAN               & \multicolumn{1}{c|}{8.40}      & \multicolumn{1}{c|}{4.04}  &   \multicolumn{1}{c|}{5.36}     &   \multicolumn{1}{c|}{10.06}   & \multicolumn{1}{c}{15.72}         \\
Restormer          & \multicolumn{1}{c|}{10.00}      &  \multicolumn{1}{c|}{6.79} &   \multicolumn{1}{c|}{\underline{4.40}}     &   \multicolumn{1}{c|}{9.78}        & \multicolumn{1}{c}{11.60}        \\
SPECAT-L          & \multicolumn{1}{c|}{4.59}      &  \multicolumn{1}{c|}{1.46} &   \multicolumn{1}{c|}{4.11}     &   \multicolumn{1}{c|}{10.09}        & \multicolumn{1}{c}{12.98}        \\
MST++              & \multicolumn{1}{c|}{6.18}      & \multicolumn{1}{c|}{{1.62}}  &  \multicolumn{1}{c|}{4.46}     & \multicolumn{1}{c|}{\underline{9.21}}  & \multicolumn{1}{c}{\underline{11.40}}         \\ \hline
UnGUN              & \multicolumn{1}{c|}{\textbf{0.15}}      & \multicolumn{1}{c|}{\textbf{0.15}}  &  \multicolumn{1}{c|}{4.17}     & \multicolumn{1}{c|}{11.05}  & \multicolumn{1}{c}{11.52}         \\ \hline
Ours               & \multicolumn{1}{c|}{\underline{4.54}}      &\multicolumn{1}{c|}{\underline{1.35}}      & \multicolumn{1}{c|}{\textbf{3.97}}     &\multicolumn{1}{c|}{\textbf{8.28}}               & \multicolumn{1}{c}{\textbf{9.65}}            \\ \hline
\end{tabular}}
\label{table3.1}
\end{table}
\renewcommand\arraystretch{1.0}

In the realm of hyperspectral tasks, KANs are capable of more naturally learning intricate high-dimensional mapping through the utilization of B-Spline. To further augment the fitting capacity of KANs for the reconstruction of diverse hyperspectral objects, a non-uniform control point matrix is incorporated.
On the NTIRE 2020 Clean Track, as presented in Table \ref{tabel1} and Table \ref{table3.1}, our proposed method outperforms sub-optimal approaches across all evaluation metrics. Specifically, it yields a reduction of 0.0050 in the RMSE, a reduction of 0.0047 in the MRAE, an improvement of 1.31 in the PSNR, and an improvement of 0.0172 in the SSIM. Additionally, in the comparative experiments, the number of parameters of our method ranks second.

Furthermore, when contrasted with other UnHIG frameworks, our method exhibits substantial enhancements on the NTIRE 2022 dataset, which is devoid of additional priors. This superiority can be attributed to our approach's consideration of both spectral and geometric properties to guide the alignment process.
\subsubsection{Ablation Analysis}
Ablation analysis is conducted by the setting in Table \ref{table3.1}. In order to demonstrate the value of Nukes compared with the traditional KANs, we design the ablation experiment about the Non-Uniform KANs in Table \ref{table5}.

\renewcommand\arraystretch{1.0}
\begin{table}[ht]
\caption{The ablation experimental results of NTIRE 2022 dataset. The best and second best results are \textbf{highlighted} and \underline{underlined}.}
\centering
\begin{tabular}{c|c|cccc}
\hline
\multirow{2}{*}{\begin{tabular}[c]{@{}c@{}}Abla-\\ tion\end{tabular}} & \multirow{2}{*}{\begin{tabular}[c]{@{}c@{}}Params\\ train(M)\end{tabular}} & \multicolumn{4}{c}{NTIRE 2022}                                                                                                    \\ \cline{3-6} 
                           &&\multicolumn{1}{c|}{RMSE}            & \multicolumn{1}{c|}{MRAE}            & \multicolumn{1}{c|}{PSNR}           & SSIM          \\ \hline
Base & \multicolumn{1}{c|}{{6.01}}                  & \multicolumn{1}{c|}{0.0742}           & \multicolumn{1}{c|}{0.8001}           & \multicolumn{1}{c|}{25.80}          & 0.7653         \\ \hline
Nukes & \multicolumn{1}{c|}{{5.87}}                         & \multicolumn{1}{c|}{0.0723}            & \multicolumn{1}{c|}{0.7996}           & \multicolumn{1}{c|}{25.99}          & 0.7815         \\
G-MSA & \multicolumn{1}{c|}{\textbf{4.42}}                         & \multicolumn{1}{c|}{\underline{0.0574}}            & \multicolumn{1}{c|}{\underline{0.5139}}           & \multicolumn{1}{c|}{\underline{27.09}}           & 0.8334         \\
DCPM-G & \multicolumn{1}{c|}{{4.54}}                         & \multicolumn{1}{c|}{0.0634}            & \multicolumn{1}{c|}{0.5816}            & \multicolumn{1}{c|}{26.93}          & \underline{0.8357}         \\
DCPM-S & \multicolumn{1}{c|}{\underline{4.53}}                         & \multicolumn{1}{c|}{{0.0623}}            & \multicolumn{1}{c|}{{0.6027}}            & \multicolumn{1}{c|}{{25.83}}           & {0.8174}         \\ \hline
Ours & \multicolumn{1}{c|}{4.54}                      & \multicolumn{1}{c|}{\textbf{0.0528}} & \multicolumn{1}{c|}{\textbf{0.4829}} & \multicolumn{1}{c|}{\textbf{27.17}} & \textbf{0.8526} \\ \hline
\end{tabular}
\label{table4}
\end{table}
As demonstrated in Table \ref{table4}, $\textbf{Ours}$ achieved the best reconstruction performance. To gain a more in-depth understanding of the role of each module, the two branches of DCPM are removed in \textbf{DCPM-Geometry} and \textbf{DCPM-Spectrum} to analyse their influence for reconstruction. When examining the results, we observe the following changes upon the removal of specific components from the model: With the removal of DCPM-Geometry (DCPM-G), both RMSE and MRAE increase by 0.0106 and 0.0987 respectively. Similarly, upon the removal of DCPM-Spectrum (DCPM-S), we again notice an increase of 0.0095 in RMSE and 0.1198 in MRAE.

In terms of parameter during training, it might seem at first glance that Ours possesses more parameters than the current leading data-driven HIG methods \cite{zamir2022restormer,li2020adaptive,cai2022mst++}. However, during the testing phase, NukesFormers relies solely on the $\mathcal{G}_{R\rightarrow H}$. Importantly, $\mathcal{L}_{adv}$ is also not used during testing, which significantly reduces its parameter count to 1.35M.

In contrast to existing KANs, Nukes meticulously takes into account the characteristic bands of hyperspectral targets through the embedding of dynamically adjustable control-point matrices. The control points introduced by the NCPG effectively steer the direction of attention in diverse scenarios. Moreover, this empowers the high-dimensional mapping approximated by Nukes to extract more distinctive and valuable features across different scenes and channels, while incurring only a slight increase in the number of parameters.

\begin{table}[ht]
\caption{The ablation experimental results about the Nukes and KANs. The best and second best results are \textbf{highlighted} and \underline{underlined}.}
\centering
\begin{tabular}{c|c|cc|cc}
\hline
\multirow{2}{*}{Model} & \multirow{2}{*}{\begin{tabular}[c]{@{}c@{}}Params\\ train(M)\end{tabular}} & \multicolumn{2}{c|}{NTIRE 2022} & \multicolumn{2}{c}{CAVE} \\
\cline{3-6} & & \multicolumn{1}{c|}{MRAE} & \multicolumn{1}{c|}{PSNR} & \multicolumn{1}{c|}{RMSE} & PSNR\\ \hline
KANs & \textbf{2.66} & \multicolumn{1}{c|}{0.7882} & 24.07 & \multicolumn{1}{c|}{0.0544} & 26.45 \\ \hline
Nukes & \underline{2.91} & \multicolumn{1}{c|}{0.7523} & 25.61 & \multicolumn{1}{c|}{0.0512} & 27.03 \\ \hline
\begin{tabular}[c]{@{}c@{}}Ours-\\ KANs\end{tabular} & 4.27 & \multicolumn{1}{c|}{\underline{0.6452}} & \underline{26.65} & \multicolumn{1}{c|}{\underline{0.0414}} & \underline{27.85} \\ \hline
\begin{tabular}[c]{@{}c@{}}Ours-\\ Nukes\end{tabular} & 4.54 & \multicolumn{1}{c|}{\textbf{0.4829}} & \textbf{27.17} & \multicolumn{1}{c|}{\textbf{0.0386}} & \textbf{29.57} \\ \hline
\end{tabular}
\label{table5}
\end{table}

To further validate the enhancement of Non-Uniform's capacity for extracting hyperspectral features, particularly in the null space, we carried out two ablation experiments employing traditional KANs and Nukes. Specifically, we initially devised a straightforward Encoder-Decoder architecture that incorporates solely KANs or Nukes along with convolutional layers for upsampling and downsampling operations. This architecture effectively verified the fitting capability of KANs in high-dimensional feature mapping and the attention regulation ability of Non - Uniform across different feature channels. Subsequently, we substituted Nukes with traditional KANs. The experimental results indicated an increase of 0.1623 in the MRAE and a decrease of 0.52 in the PSNR in NTIRE 2022. This evidence demonstrates that Nukes facilitates the model in better recovering null-space details in HIG tasks.

\section{Conclusion}
In this work, we introduce NukesFormers, a novel framework for UnHIG. The proposed method leverages the RND methodology to systematically partition the complex spectral distribution of objects into range space and null space. By incorporating non-uniform KANs, the framework effectively extracts and compensates for the spectral attributions within the null space. Furthermore, our model seamlessly integrates the HSI and RGB (or MSI) domains through cross-domain contrastive learning, facilitating enhanced spectral representation and domain alignment. Comparative experiments highlight the superior performance of NukesFormers, underscoring its robustness, reliability, and consistency across diverse scenes.